\documentclass[11pt,a4paper]{article}
\usepackage[hyperref]{acl2021}
\usepackage{times}

\usepackage{subcaption}

\usepackage{graphicx}

\usepackage[ruled,vlined]{algorithm2e}
\usepackage{booktabs}
\usepackage{multirow}
\usepackage{latexsym}
\usepackage{bm}
\usepackage{pifont}
\newcommand{\cmark}{\ding{51}}%
\newcommand{\xmark}{\ding{55}}%

\usepackage{todonotes}

\usepackage{amsmath}
\DeclareMathOperator*{\argmax}{argmax}

\usepackage{microtype}

\aclfinalcopy 
\def\aclpaperid{2187} 

\title{Understanding the Properties of Minimum Bayes Risk Decoding\\ in Neural Machine Translation}

\author{Mathias M\"uller$^1$ \and Rico Sennrich$^{1,2}$\\
  $^1$Department of Computational Linguistics, University of Zurich\\
  $^2$School of Informatics, University of Edinburgh}

\date{}

\begin{document}
\maketitle
\begin{abstract}

Neural Machine Translation (NMT) currently exhibits biases such as producing translations that are too short and overgenerating frequent words, and shows poor robustness to copy noise in training data or domain shift. Recent work has tied these shortcomings to beam search -- the de facto standard inference algorithm in NMT -- and \citet{eikema-aziz-2020-map} propose to use Minimum Bayes Risk (MBR) decoding on unbiased samples instead.

In this paper, we empirically investigate the properties of MBR decoding on a number of previously reported biases and failure cases of beam search. We find that MBR still exhibits a length and token frequency bias, owing to the MT metrics used as utility functions, but that MBR also increases robustness against copy noise in the training data and domain shift.\footnote{Code and documentation available at \url{https://github.com/ZurichNLP/understanding-mbr}}

\end{abstract}

\section{Introduction}

Neural Machine Translation (NMT) currently suffers from a number of issues such as underestimating the true length of translations \citep{koehn-knowles-2017-six,stahlberg-byrne-2019-nmt,kumar2019calibration}, underestimating the probability of rare words and over-generating very frequent words \citep{ott2018analyzing}, or being susceptible to copy noise in the training data \citep{khayrallah-koehn-2018-impact}. In out-of-domain translation, \textit{hallucinations} (translations that are fluent but unrelated to the source) are common \citep{koehn-knowles-2017-six,lee2018hallucinations,muller-etal-2020-domain}.

Previous work has addressed these problems with decoding heuristics such as length normalization \citep{wu2016google}, data cleaning \citep{junczys-dowmunt-2018-dual,banon-etal-2020-paracrawl} or model regularization \citep{bengio-etal-2015-scheduled,shen-etal-2016-minimum,wiseman-rush-2016-sequence,zhang-etal-2019-bridging,ng-etal-2020-ssmba}.

Recently, \citet{eikema-aziz-2020-map} have highlighted the role of the decision rule, namely searching for the highest-scoring translation, and have argued that it is at least partially to blame for some of these biases and shortcomings.
They found that sampling from an NMT model is faithful to the training data statistics, while beam search is not. They recommend the field look into alternative inference algorithms
based on unbiased samples, such as Minimum Bayes Risk (MBR) decoding.

We believe MBR has potential to overcome several known biases of NMT. More precisely, if a bias can be understood as being caused by the mode-seeking nature of beam search then we hypothesize that MBR could exhibit less bias. We view short translations, copies of the source text and hallucinations as hypotheses that are probable, but quite different to other probable hypotheses. If such pathological hypotheses are in a pool of samples, it is unlikely that MBR would select them as the final translation.

While \citet{eikema-aziz-2020-map} compare the statistical properties of samples and beam search outputs, and show that MBR can perform favourably compared to beam search according to automatic metrics, our paper aims to perform a targeted study of MBR and its properties, specifically its effects on the biases and shortcomings discussed previously. In our experiments we find that

\begin{itemize}
    \item If used with a utility function that favours short translations, MBR inherits this bias;
    \item MBR still exhibits a token probability bias in that it underestimates the probability of rare tokens and overestimates very common tokens;
    \item Compared to beam search, MBR decoding is more robust to copy noise in the training data;
    \item MBR exhibits higher domain robustness than beam search. We demonstrate that MBR reduces the amount of hallucinated content in translations.
\end{itemize}

\section{Background}

\subsection{Maximum-a-posteriori (MAP) decoding}

The de facto standard decoding algorithm in NMT is beam search \citep{graves2012sequence,boulanger2013audio,sutskever-2014}. Beam search belongs to a broader class of inference procedures called maximum-a-posteriori (MAP) algorithms. What MAP algorithms have in common is that they attempt to find the most probable translation under a given model. Essentially, they try to recover the \textit{mode} of the output distribution over sequences.

An exact solution to this search problem is usually intractable. Beam search is an approximation that is tractable, but it also frequently fails to find the true mode of the distribution \citep{stahlberg-byrne-2019-nmt}.

\subsection{Known deficiencies of NMT systems}
\label{subsec:deficiencies}

NMT systems are known to be deficient in a number of ways. We describe here only the ones relevant to our discussion and experiments. 

\textbf{Length bias:} Systems underestimate the true length of translations. On average, their translations are shorter than references \citep{koehn-knowles-2017-six,stahlberg-byrne-2019-nmt,kumar2019calibration}.

\textbf{Skewed word frequencies:} In translations, tokens that occur frequently in the training data are overrepresented. On the other hand, rare tokens occur fewer times than their probability in the training data would suggest \citep{ott2018analyzing}.

\textbf{Beam search curse:} Increasing the beam size leads to finding translations that are more probable under the model. In theory, this should improve translation quality. Paradoxically, empirical results show that large beam sizes decrease quality \citep{koehn-knowles-2017-six,ott2018analyzing}. 

\textbf{Susceptibility to copy noise:} Copied content in the training data disproportionately affects translation quality. More specifically, the most detrimental kind are copies of the source sentence on the target side of the training data \citep{khayrallah-koehn-2018-impact}. If such copies are present in the training data, copy hypotheses will be overrepresented in beam search \citep{ott2018analyzing}.

\textbf{Low domain robustness:} Systems are not robust under distribution shifts such as domain shift. Having a system translate in an unknown test domain often does not gradually degrade translation quality, but leads to complete failure cases called hallucinations \citep{lee2018hallucinations,koehn-knowles-2017-six,muller-etal-2020-domain}.

Much past research has attributed those deficiencies to model architectures or training algorithms, while treating beam search as a fixed constant in experiments. In contrast, \citet{eikema-aziz-2020-map} argue that the fit of the model is reasonable, which means that neither the model itself nor its training can be at fault. Rather, they argue that the underlying problem is beam search.

\textbf{Inadequacy of the mode:} \citet{stahlberg-byrne-2019-nmt} and \citet{eikema-aziz-2020-map} suggest that the mode of the distribution over output sequences is in fact not the best translation. On the contrary, it seems that in many cases the mode is the empty sequence
\citep{stahlberg-byrne-2019-nmt}. In addition, it appears that the probability of the mode is not much different from very many other sequences, as the output distribution is quite flat in an extensive region of output space \citep{eikema-aziz-2020-map}.

Intuitively, it makes sense that such a situation could arise in NMT training: maximum likelihood estimation training does not constrain a model to be characterized well by its mode only. If the mode is inadequate, then obviously that is problematic for a mode-seeking procedure such as beam search, and MAP inference in general. In fact, MAP decoding should be used only if the mode of the output distribution can be trusted \citep{smith:2011:synthesis}.

An alternative is a decision rule that considers how different a translation is from other likely translations.

\subsection{Minimum Bayes Risk Decoding}

MBR decoding was used in speech recognition \citep{goel-byrne-2000} and statistical machine translation \citep{kumar-byrne-2004-minimum,tromble-etal-2008-lattice}. More recently, MBR was also used to improve beam search decoding in NMT \citep{stahlberg-etal-2017-neural,shu2017later,blain2017exploring}. \citet{eikema-aziz-2020-map} are the first to test a variant of MBR that operates on samples instead of an nbest list generated by beam search.

We give here a simplified, accessible definition of MBR in the context of NMT. Essentially, the goal of MBR is to find not the most probable translation, but the one that minimizes the expected risk for a given loss function and the true posterior distribution.
In practice, the set of all possible candidate translations can be approximated by drawing from the model a pool of samples $\mathcal{S}$ of size $n$:

\begin{equation}
\mathcal{S} = (s_1, ..., s_n) \sim p(y|x,\theta).
\end{equation}

The same set of samples can also be used to approximate the true posterior distribution. Then for each sample $s_i$ in $\mathcal{S}$, its expected utility (the inverse risk) is computed by comparing it to all other samples in the pool. The sample with the highest expected utility is selected as the final translation:

\begin{equation}
y^\star = \argmax_{s_i \in \mathcal{S}} \frac{1}{n} \sum_{s_j = 1}^n u(s_i, s_j)
\end{equation}

The size of the pool $n$ and the utility function $u$ are hyperparameters of the algorithm. A particular utility function typically computes the \textit{similarity} between a hypothesis and a reference translation. Therefore, MBR ``can be thought of as selecting a \textit{consensus} translation [...] that is closest on average to all likely translations'' \citep{kumar-byrne-2004-minimum}.

\section{Motivation for experiments}

We hypothesize that MBR decoding is useful for a certain class of failure cases encountered with beam search. Namely, if an incorrect translation from beam search can be characterized as a hypothesis that is likely but fairly different from other hypotheses with similar probability, then MBR is expected to improve over beam search.

Several known deficiencies of NMT systems outlined in Section \ref{subsec:deficiencies} belong to this class of beam search failures. For instance, length bias occurs when a beam search translation is shorter than other hypotheses with comparable probability. Likewise, translations that are copies of the input sentence or hallucinations (translations that are fluent, but unrelated to the input) can be avoided with MBR if they are not common in a pool of samples.

Finally, we study the skewedness of token frequencies in translations. \citet{eikema-aziz-2020-map} study lexical biases in NMT models, showing that model samples have higher agreement with the training distribution than MAP output. We investigate whether this is also true for MBR decoding, focusing on the well-known bias towards frequent tokens.

\section{Experimental Setup}

\subsection{Data}

We use data for a number of language pairs from the Tatoeba Challenge \citep{tiedemann:2020:WMT}. Individual language pairs are fairly different in terms of language families, scripts and training set sizes. See Appendix \ref{sec:data-set-details} for details about our data sets.

For one additional experiment on out-of-domain robustness we use data from \citet{muller-etal-2020-domain}. This data set is German-English and defines 5 different domains of text (medical, it, koran, law and subtitles). Following \citet{muller-etal-2020-domain} we train our model on the medical domain, and use data in other domains to test domain robustness.

We hold out a random sample of the training data for testing purposes. The size of this sample varies between 1k and 5k sentences, depending on the overall size of the training data.

\subsection{Models}

Our preprocessing and model settings are inspired by OPUS-MT \citep{TiedemannThottingal:EAMT2020}. We use Sentencepiece \citep{kudo-2018-subword} with subword regularization as the only preprocessing step, which takes care of both tokenization and subword segmentation. The desired number of pieces in the vocabulary varies with the size of the data set.

We train NMT models with Sockeye 2 \citep{domhan-etal-2020-sockeye}. The models are standard Transformer models \citep{DBLP:journals/corr/VaswaniSPUJGKP17}, except that some settings (such as word batch size and dropout rate) vary with the size of the training set. Following \citet{eikema-aziz-2020-map} we disable label smoothing so as to get unbiased samples.

\begin{figure*}
    \centering
    \includegraphics[width=\textwidth]{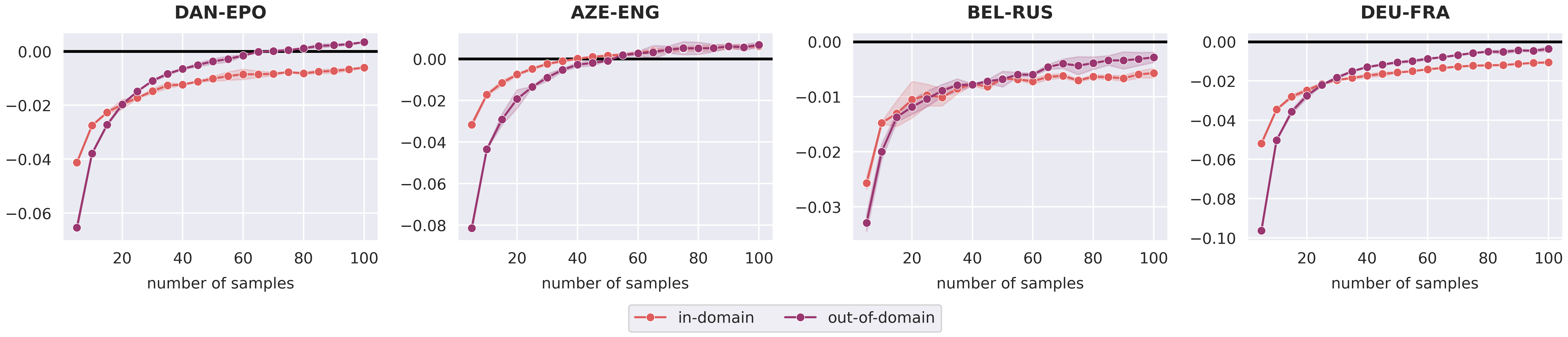}
    \caption{CHRF1 scores of MBR decoding on two test corpora: the standard Tatoeba test set (out-of-domain) and a test set of held-out training data (in-domain). Plots show the \textbf{difference} between MBR and beam search, as a function of the number of samples used for MBR.}
    \label{fig:results_compare_corpora}
\end{figure*}

\subsection{Decoding and evaluation}

In all experiments, we compare beam search to MBR decoding and in most cases also to single samples. For beam search, we always use a beam size of 5. Single samples are drawn at least 100 times to show the resulting variance.

If not stated otherwise, all results presented are on a test set held out from the training data, i.e. are certainly in-domain, which avoids any unintended out-of-domain effects.

We evaluate automatic translation quality with BLEU \citep{papineni-etal-2002-bleu}, CHRF \citep{popovic-2016-chrf} and METEOR \citep{denkowski-lavie-2014-meteor}. We compute BLEU and CHRF with SacreBLEU \citep{post-2018-call}. See Appendix \ref{sec:evaluation-details} for details.

MBR also depends on samples, so we repeat each MBR experiment twice to show the resulting variance. We also vary the number of samples used with MBR, from 5 to 100 in increments of 5. Finally, we produce MBR translations with different utility functions. All of the utility functions are sentence-level variants of our evaluation metrics: BLEU,
CHRF or METEOR. See Table \ref{tab:utility-functions} for an overview of utility functions.
If not stated otherwise, MBR results are based on 100 samples and use \texttt{chrf-1} as the utility function.

\begin{table}
    \small
    \centering
    \begin{tabular}{llcccc}
      \toprule
       & \multicolumn{1}{c}{\textbf{smoothed?}} & \multicolumn{1}{c}{\bm{$\alpha$}} & \multicolumn{1}{c}{\bm{$\beta$}} & \multicolumn{1}{c}{\bm{$\gamma$}} & \multicolumn{1}{c}{\bm{$\delta$}} \\
      \midrule
      bleu & \xmark & - & - & - & - \\
      bleu-floor & \cmark & - & - & - & - \\
      bleu-add-k & \cmark & - & - & - & - \\
      bleu-exp & \cmark & - & - & - & - \\
      \midrule
      chrf-0.5 & \xmark & - & 0.5 & - & - \\
      chrf-1 & \xmark & - & 1.0 & - & - \\
      chrf-2 & \xmark & - & 2.0 & - & - \\
      chrf-3 & \xmark & - & 3.0 & - & - \\
      \midrule
      meteor & \xmark & 0.85 & 0.2 & 0.6 & 0.75 \\
      meteor-0.5 & \xmark & 0.50 & 0.2 & 0.6 & 0.75 \\
      \bottomrule
    \end{tabular}
    \caption{Utility functions used with MBR. The smoothed variants of BLEU correspond to the ones implemented in SacreBLEU \citep{post-2018-call} and are defined in \citet{chen-cherry-2014-systematic}.}
    \label{tab:utility-functions}
\end{table}

\section{Length bias}
\label{sec:length-bias}

\begin{figure*}
    \centering
    \includegraphics[width=\textwidth]{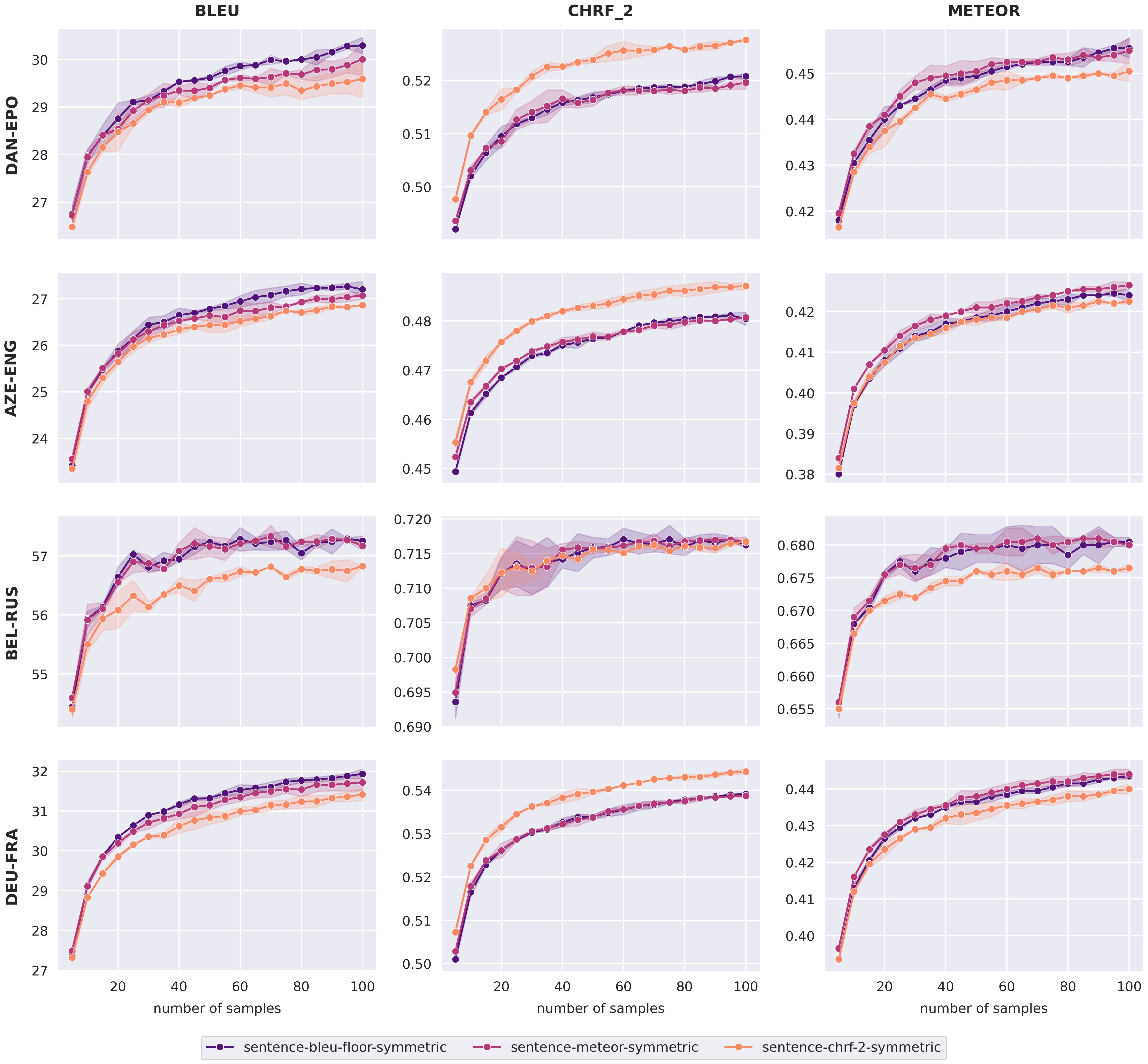}
    \caption{Comparison of MBR utility functions. Different columns show translation quality as measured by a particular evaluation metric. Line colors refer to different utility functions. Shaded areas show standard deviation.}
    \label{fig:results_compare_risk_functions}
\end{figure*}

We evaluate MBR decoding with different utility functions. There is no single utility function which performs best on all evaluation metrics. Instead, any of our evaluation metrics can be optimized by choosing a closely related utility function (see Figure \ref{fig:results_compare_risk_functions} and Appendix \ref{sec:additional-comparisons-between}). For instance, \texttt{chrf-2} as the utility function leads to the best CHRF2 evaluation scores.

\textbf{Number of samples:} We find that the translation quality of MBR increases steadily as the number of samples grows (see Figure \ref{fig:results_compare_risk_functions}). This means that MBR does not suffer from the beam search curse where single pathological hypotheses in a large beam can jeopardize translation quality.


\begin{table*}
    \centering
    \begin{tabular}{lcccc}
      \toprule
       & \multicolumn{1}{c}{\textbf{DAN-EPO}} & \multicolumn{1}{c}{\textbf{AZE-ENG}} & \multicolumn{1}{c}{\textbf{BEL-RUS}} & \multicolumn{1}{c}{\textbf{DEU-FRA}} \\
      \midrule
      reference & 11.91 & 15.54 & 8.41 & 20.19 \\
      \midrule
      sample & 11.73 & 15.15 & 8.29 & 19.99 \\
      beam-normalized & 11.61 & 14.45 & 8.23 & 19.62 \\
      beam-unnormalized & 11.21 & 13.62 & 8.20 & 19.08 \\
      \midrule
      bleu-floor & 11.51 & 14.41 & 8.18 & 19.55 \\
      meteor & 12.23 & 15.29 & 8.26 & 20.38 \\
      chrf-2 & 12.50 & 15.88 & 8.31 & 20.89 \\
      \midrule
      bleu-floor-symmetric & 11.51 & 14.34 & 8.19 & 19.53 \\
      meteor-symmetric & 11.47 & 14.12 & 8.20 & 19.40 \\
      chrf-2-symmetric & 11.48 & 14.16 & 8.18 & 19.40 \\
      \midrule
      chrf-0.5 & 10.63 & 12.99 & 8.08 & 18.02 \\
      \bottomrule
    \end{tabular}
    \caption{Lengths of hypotheses as mean number of tokens.}
    \label{tab:lengths}
\end{table*}

\begin{figure*}
    \centering
    \includegraphics[width=\textwidth]{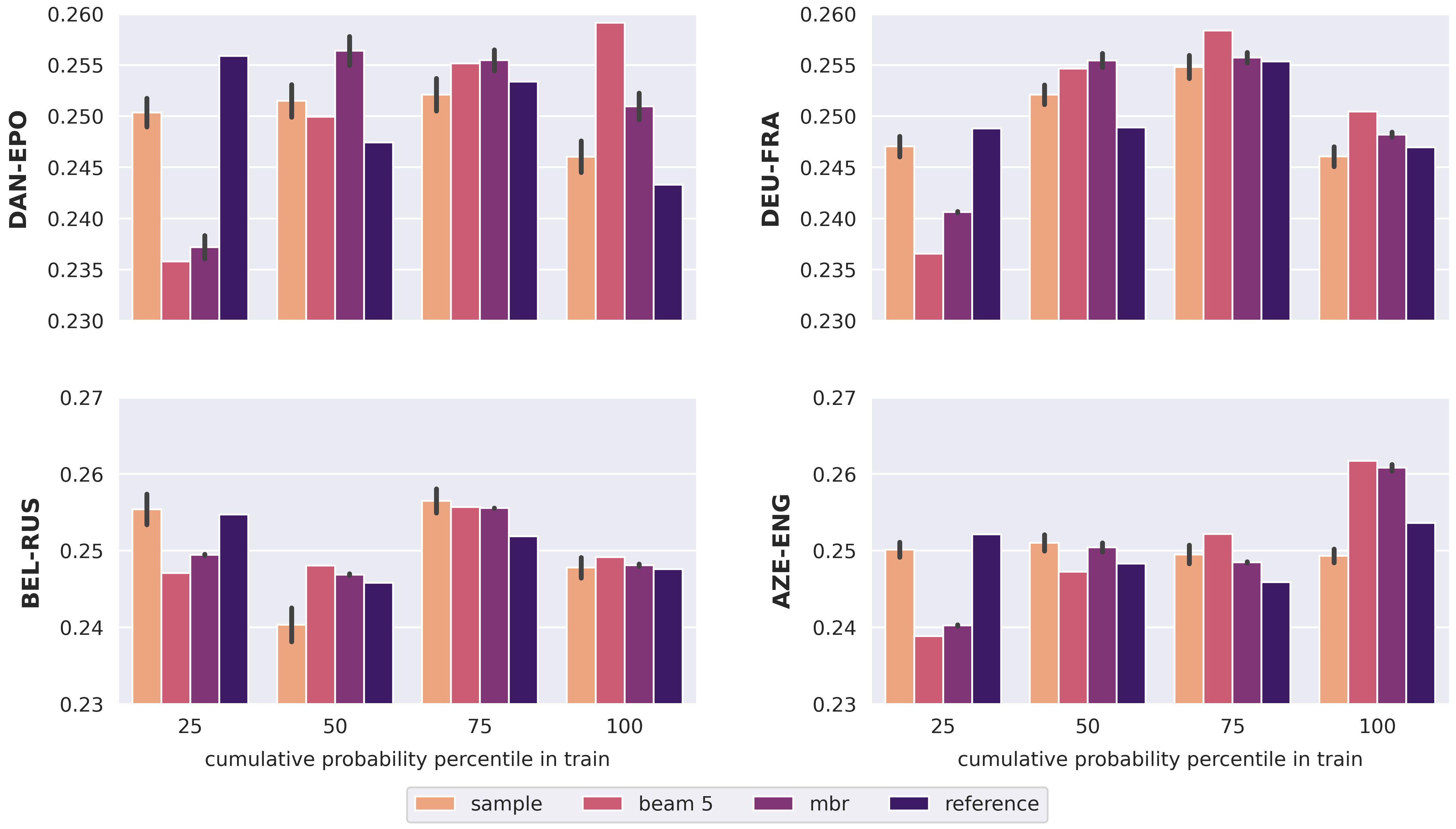}
    \caption{Probability of tokens in translations (x-axis) bucketed by frequency in training data (y-axis). Vertical bars indicate standard deviation for methods that involve sampling.}
    \label{fig:results_count_analysis}
\end{figure*}

We analyze the lengths of translations produced by different decoding methods in Table \ref{tab:lengths} (see Appendix \ref{sec:additional-length-tables} for additional statistics). We find that in terms of mean length of translations, beam search underestimates the true length of translations, even when hypotheses are normalized. Hypotheses generated by sampling better match the reference length.
This is in line with the findings of \citet{eikema-aziz-2020-map}.

For MBR decoding, it is clear that the choice of utility function has an impact on the mean length of the resulting translations. For instance, employing sentence-level BLEU as the utility function leads to translations that are too short. BLEU is a precision-based metric known to prefer shorter translations on the sentence level \citep{nakov-etal-2012-optimizing}.

\texttt{chrf-2} and \texttt{meteor} emphasize recall more, and the resulting MBR translations overestimate the true length of translations.\footnote{While \citet{popovic-2016-chrf} find that the recall-biased CHRF2 achieves the highest correlation with human judgments as an evaluation metric, this does not entail that the same recall bias is optimal in the utility function for MBR.} On the other hand, \texttt{chrf-0.5}, a CHRF variant with a bias for precision, leads to the shortest translations overall.

We test whether we can reduce length biases by symmetrizing our utility functions $u$ as follows:
\begin{equation}
u_{sym}(s_i, s_j) = H(u(s_i, s_j), u(s_j, s_i))
\end{equation}
\noindent where $H$ is the harmonic mean. This should avoid favouring either recall or precision, but in practice even symmetric utility functions lead to translations that are shorter than references on average. 

Based on these observations we conclude that \textbf{MBR inherits length biases associated with its utility function}. 

\section{Token frequency bias}
\label{sec:token-frequency-bias}

Beam search overgenerates tokens that are very common in the training data and undergenerates rare tokens (see Section \ref{subsec:deficiencies}). Sampling on the other hand assigns correct probabilities to common and rare tokens. Given that MBR is based on samples, does it share this property with sampling?

In Figure \ref{fig:results_count_analysis} we show that this is not the case. Although the skewedness of probabilities is less severe for MBR than for beam search, MBR still assigns too high a probability to frequent events. 
A reason for this is that our utility functions are based on surface similarity between samples, so rare tokens, which will be sampled rarely, will thus also have low utility.

Unfortunately, there is a \textbf{trade-off between correct probability statistics for very common and very rare words and translation quality}. The most faithful statistics can be obtained from sampling, but sampling leads to the worst overall translation quality.

\section{Domain robustness}
\label{sec:domain-robustness}

\begin{figure*}
    \centering
    \includegraphics[width=\textwidth]{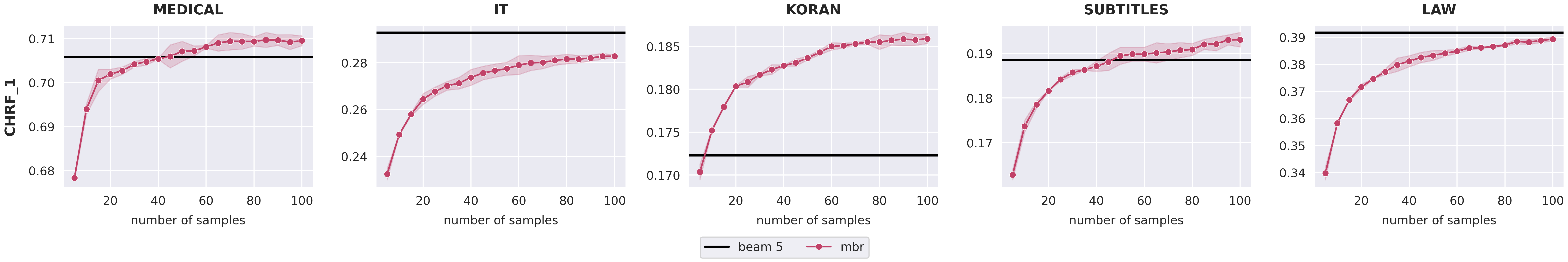}
    \caption{CHRF1 scores of MBR and beam search on the domain robustness benchmark of \citet{muller-etal-2020-domain}. The \textit{medical} test set is in-domain, the remaining sets are out-of-domain.}
    \label{fig:results_domain_robustness}
\end{figure*}

\begin{figure*}
    \centering
    \includegraphics[width=\textwidth]{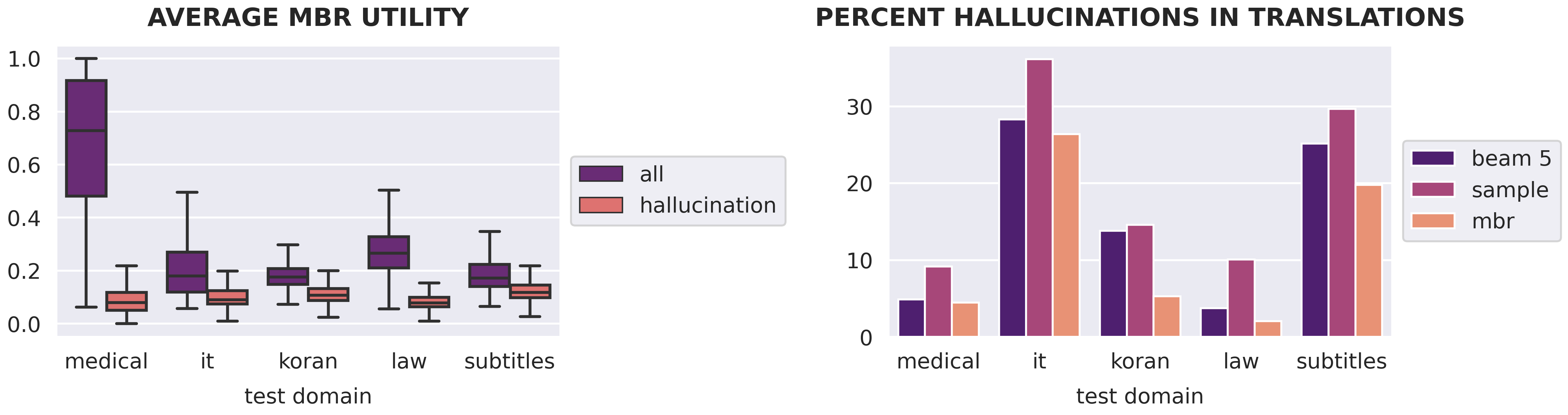}
    \caption{Analysis of hallucinations in MBR and beam translations. Left: Average utility of hallucination hypotheses in pools of samples. Right: how often hallucinations occur in final translations.}
    \label{fig:results_hallucination_analysis}
\end{figure*}

In general, as the number of samples grows, MBR approaches but does not outperform beam search on our in-domain data (see Figure \ref{fig:results_compare_corpora}). On our out-of-domain data, the gap between MBR and beam search is smaller. We hypothesize that MBR may be useful for out-of-domain translation.

We evaluate MBR on a domain robustness benchmark by \citet{muller-etal-2020-domain}. Figure \ref{fig:results_domain_robustness} shows that on this benchmark MBR outperforms beam search on 2 out of 4 unknown test domains. A possible reason why MBR is able to outperform beam search in unknown domains is that it reduces hallucinated translations. To test this hypothesis, we define a \textit{hallucination} as a translation that has a CHRF2 score of less than $0.01$ when compared to the reference, inspired by \citet{lee2018hallucinations}.

Given this definition of hallucination, Figure \ref{fig:results_hallucination_analysis} shows that on average, MBR assigns a lower utility score to hypotheses that are hallucinations. Similarly, MBR reduces the percentage of hallucinations found in the final translations, compared to beam search or sampling. To summarize, we find that \textbf{MBR decoding has a higher domain robustness than beam search}.

\begin{figure*}
    \centering
    \includegraphics[width=\textwidth]{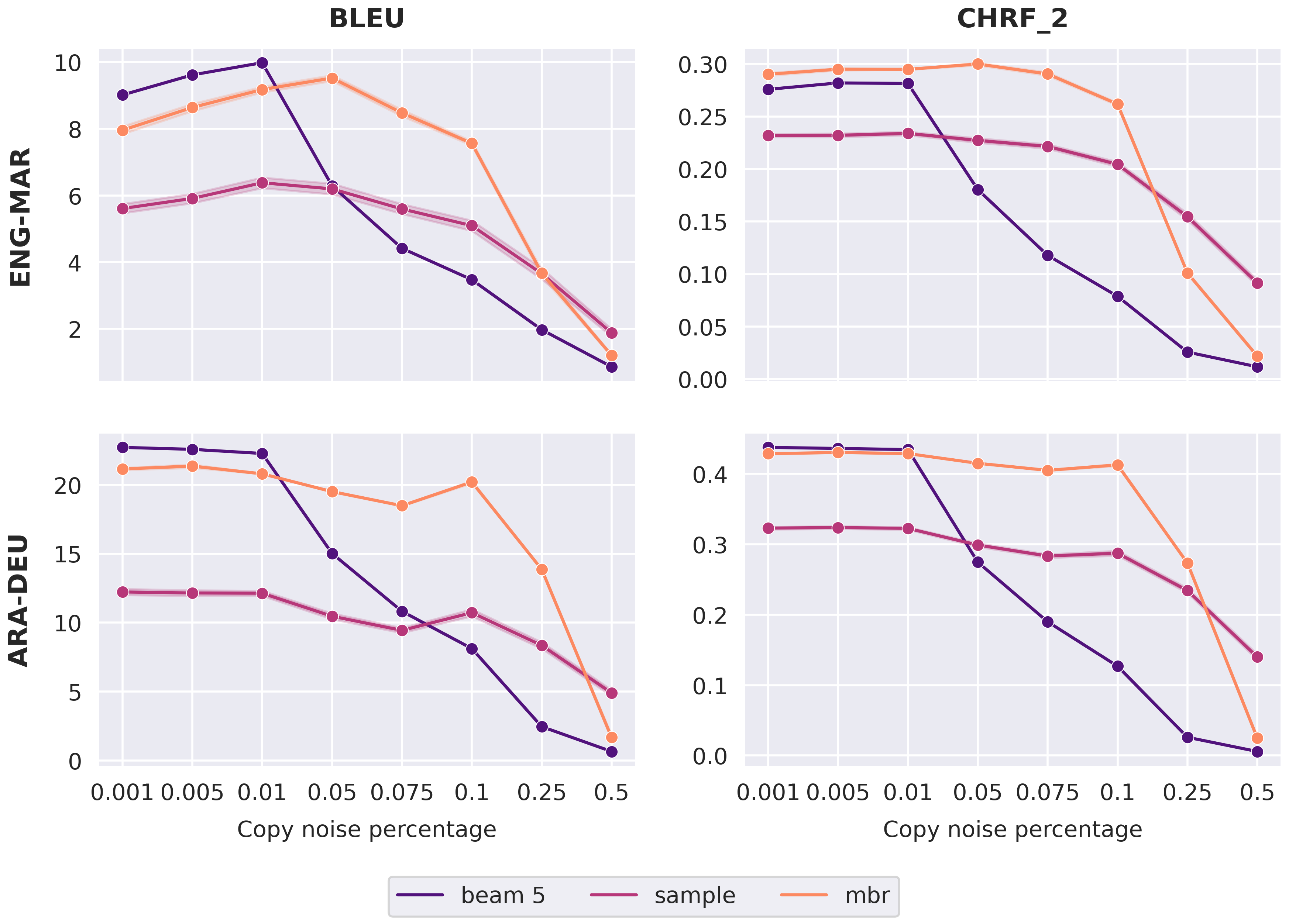}
    \caption{Susceptibility to copy noise in training data.}
    \label{fig:results_copy_noise}
\end{figure*}

\begin{figure*}
    \centering
    \includegraphics[width=\textwidth]{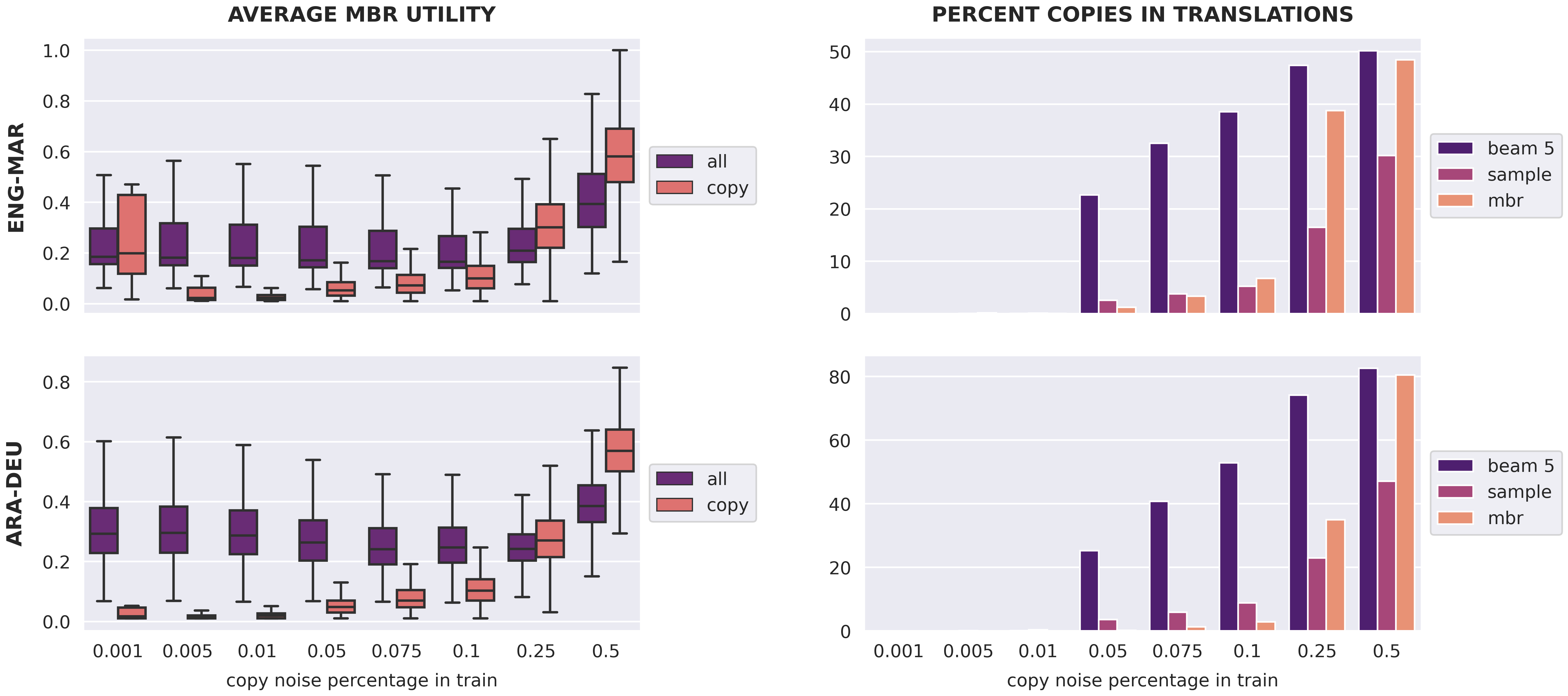}
    \caption{Analysis of copies in MBR and beam translations. Left: Average utility of copy hypotheses in pools of samples. Right: how often copies occur in final translations.}
    \label{fig:results_copy_analysis}
\end{figure*}

\section{Impact of copy noise in the training data}
\label{sec:impact-of-copy-noise-in-the-training-data}

If copies of source sentences are present on the target side of training data, copies are overrepresented in beam search (Section \ref{subsec:deficiencies}). Here we test whether MBR suffers from this copy bias as well.

We create several versions of our training sets where source copy noise is introduced with a probability between 0.1\% and 50\%. As shown in Figure \ref{fig:results_copy_noise}, MBR and beam search are comparable if there are few copies in the training data. However, if between 5 and 25\% of all training examples are copies, then MBR outperforms beam search by a large margin ($>10$ BLEU for Arabic-German).

As further evidence for the ability of MBR to tolerate copy noise we present an analysis of copies in Figure \ref{fig:results_copy_analysis}. We define a \textit{copy} as a translation with a word overlap with the reference of more than $0.9$. We show that MBR assigns a much lower utility to copy hypotheses than to all hypotheses taken together.
In the final translations, MBR manages to reduce copies substantially. For instance, if around 10\% of the training examples are copies, beam search produces around 50\% copies, while MBR reduces this number to below 10\%.

We conclude from this experiment that \textbf{MBR is more robust to copy noise in the training data}. We acknowledge that this setting is artificial because copy noise can easily be removed from data sets. Nonetheless, it is a striking example of a known shortcoming of NMT systems usually attributed to the model or training procedure, when in fact beam search is at least partially to blame.

\section{Conclusion and future work}

MBR decoding has recently regained attention in MT as a decision rule with the potential to overcome some of the biases of MAP decoding in NMT.
We empirically study the properties of MBR decoding with common MT metrics as utility functions, and find it still exhibits a length bias and token frequency bias similar to beam search. The length bias is closely tied to the utility function.
However, we also observe that MBR decoding successfully mitigates a number of well-known failure modes of NMT, such as spurious copying, or hallucinations under domain shift. The mechanism by which MBR achieves such robustness is that copies or hallucinated hypotheses in a pool of samples are assigned low utility and never selected as the final translation.

In our experiments, MBR did not generally outperform beam search according to automatic metrics, but we still deem it a promising alternative to MAP decoding due to its robustness. For future work, we are interested in exploring more sophisticated similarity metrics to be used as utility functions, including trainable metrics such as COMET \citep{rei-etal-2020-comet}, and investigating how these utility functions affect the overall quality and biases of translations.

\section{Note on reproducibility}

We will not only release the source code used to train our models (as is common in NLP papers at the moment), but
a complete pipeline of code that can be run on any instance in a fully automated fashion. This will allow to reproduce
our results, including the graphs and tables shown in this paper, in a consistent way with minimal changes. We encourage the community to attempt to reproduce our results and publish the results.

\clearpage

\section*{Acknowledgements}

This work has received funding from the Swiss National Science Foundation (grant numbers \texttt{105212-169888} and \texttt{176727}). Also, we have been assisted by the computing services of the University of Zurich (S3IT).

We would like to thank Bryan Eikema for his help with our implementation of MBR. We also thank Jörg Tiedemann, Annette Rios and Tannon Kew for helpful comments and discussion.

\bibliographystyle{acl_natbib}
\bibliography{bibliography}

\clearpage

\appendix
\onecolumn


\section{Data set details}
\label{sec:data-set-details}

\begin{table*}[h]
    \centering
    \begin{tabular}{llrl}
      \toprule
       \multicolumn{1}{c}{\textbf{ISO3 abbreviation}} & \multicolumn{1}{l}{\textbf{language pair}} & \multicolumn{1}{r}{\textbf{size}} & \multicolumn{1}{l}{\textbf{scripts}} \\
      \midrule
      DAN-EPO & Danish-Esperanto & 110k & Roman-Roman \\
      AZE-ENG & Azerbaijani-English & 680k & Roman$^{\star}$-Roman \\
      BEL-RUS & Belarusian-Russian & 70k & Cyrillic-Cyrillic \\
      DEU-FRA & German-French & 47m & Roman-Roman \\
      \midrule
      ENG-MAR & English-Marathi & 370k & Roman-Devanagari \\
      ARA-DEU & Arabic-German & 12m & Arabic-Roman \\
      \midrule
      DEU-ENG & German-English & 1m & Roman-Roman \\
      \bottomrule
    \end{tabular}
    \caption{Details about data sets. Size refers to the number of sentence pairs in the training data. Roman$^{\star}$ = Roman script with some modifications.}
    \label{tab:data-set-details}
\end{table*}

\section{Evaluation details}
\label{sec:evaluation-details}

For evaluation metrics that require tokenization (BLEU and METEOR), we use the standard \texttt{mteval13a} tokenization implemented in SacreBLEU. We do not use any language-specific tokenization rules even if they are available for the target language. The SacreBLEU signatures for our CHRF and BLEU evaluation metrics are listed in Table \ref{tab:signatures}.

\begin{table*}[h]
    \centering
    \begin{tabular}{ll}
      \toprule
       \multicolumn{1}{l}{\textbf{evaluation metric}} & \multicolumn{1}{l}{\textbf{SacreBLEU signature}} \\
      \midrule
      CHRF\_1 & chrF1+numchars.6+space.false+version.1.4.14 \\
      CHRF\_2 & chrF2+numchars.6+space.false+version.1.4.14 \\
      CHRF\_3 & chrF3+numchars.6+space.false+version.1.4.14 \\
      \midrule
      BLEU & BLEU+case.mixed+numrefs.1+smooth.exp+tok.13a+version.1.4.14 \\
      \bottomrule
    \end{tabular}
    \caption{SacreBLEU signatures of evaluation metrics.}
    \label{tab:signatures}
\end{table*}

\section{Comments on the development sets distributed with the Tatoeba challenge}
\label{sec:comments-on-dev-sets}

The Tatoeba Challenge \citep{tiedemann:2020:WMT} distributes training, development and test data for a large number of language pairs. What is peculiar about the challenge is that the training data is assembled from various sources through OPUS \citep{tiedemann-2012-parallel}, while the development and test data are contributed by users of Tatoeba\footnote{\url{https://tatoeba.org}}. This means that the development and test set can be considered out-of-domain material.

We investigated this issue and conclude that it does not constitute a problem. When both the development and test data are sampled from the training data, the results are similar to the ones we present in this paper, except for a small overall shift.

\section{Additional comparisons between utility functions}
\label{sec:additional-comparisons-between}

\begin{figure*}
    \centering
    \includegraphics[width=\textwidth]{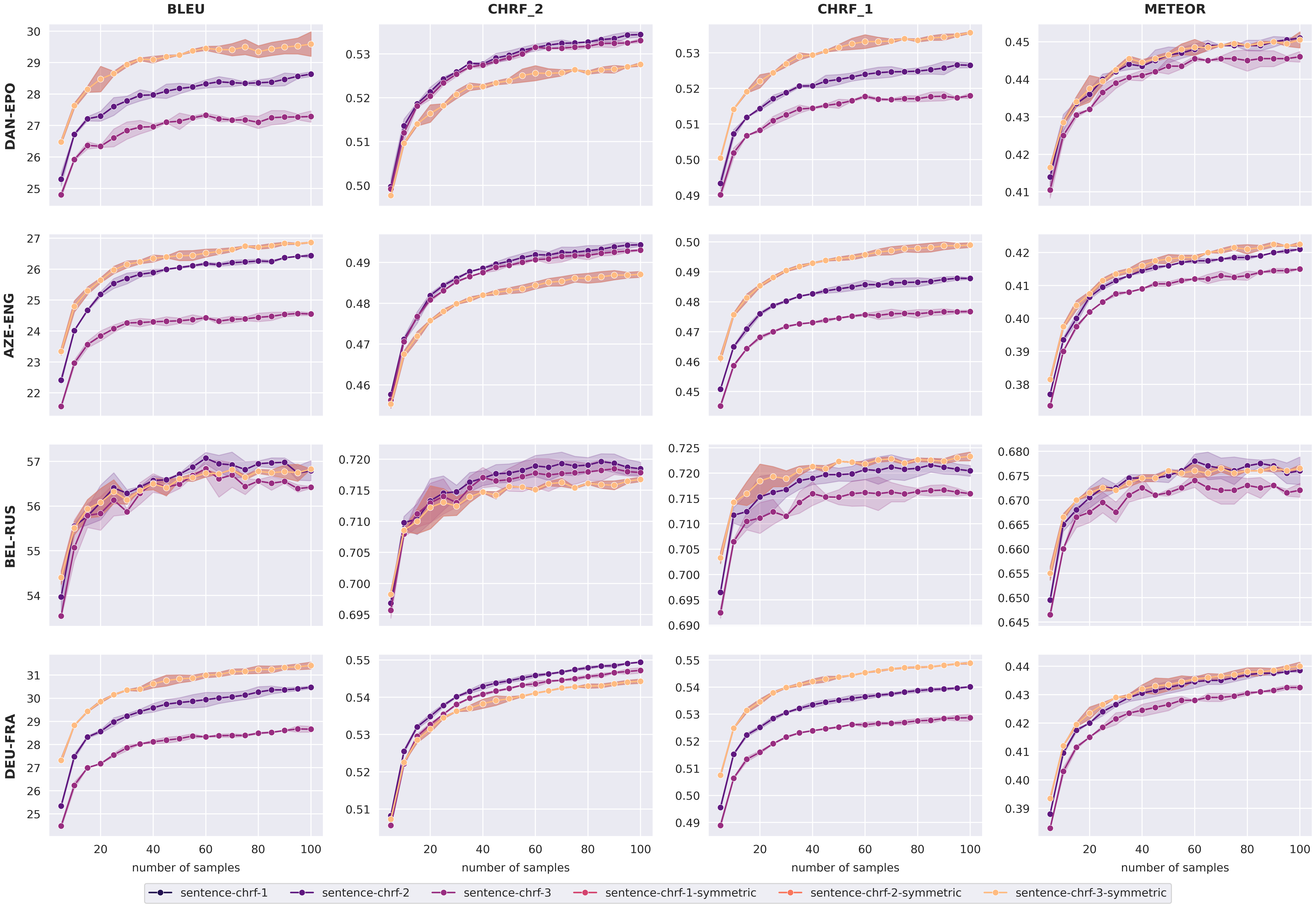}
    \caption{Comparison of utility functions that are variants of CHRF.}
    \label{fig:results_compare_risk_functions_chrf}
\end{figure*}

\begin{figure*}
    \centering
    \includegraphics[width=\textwidth]{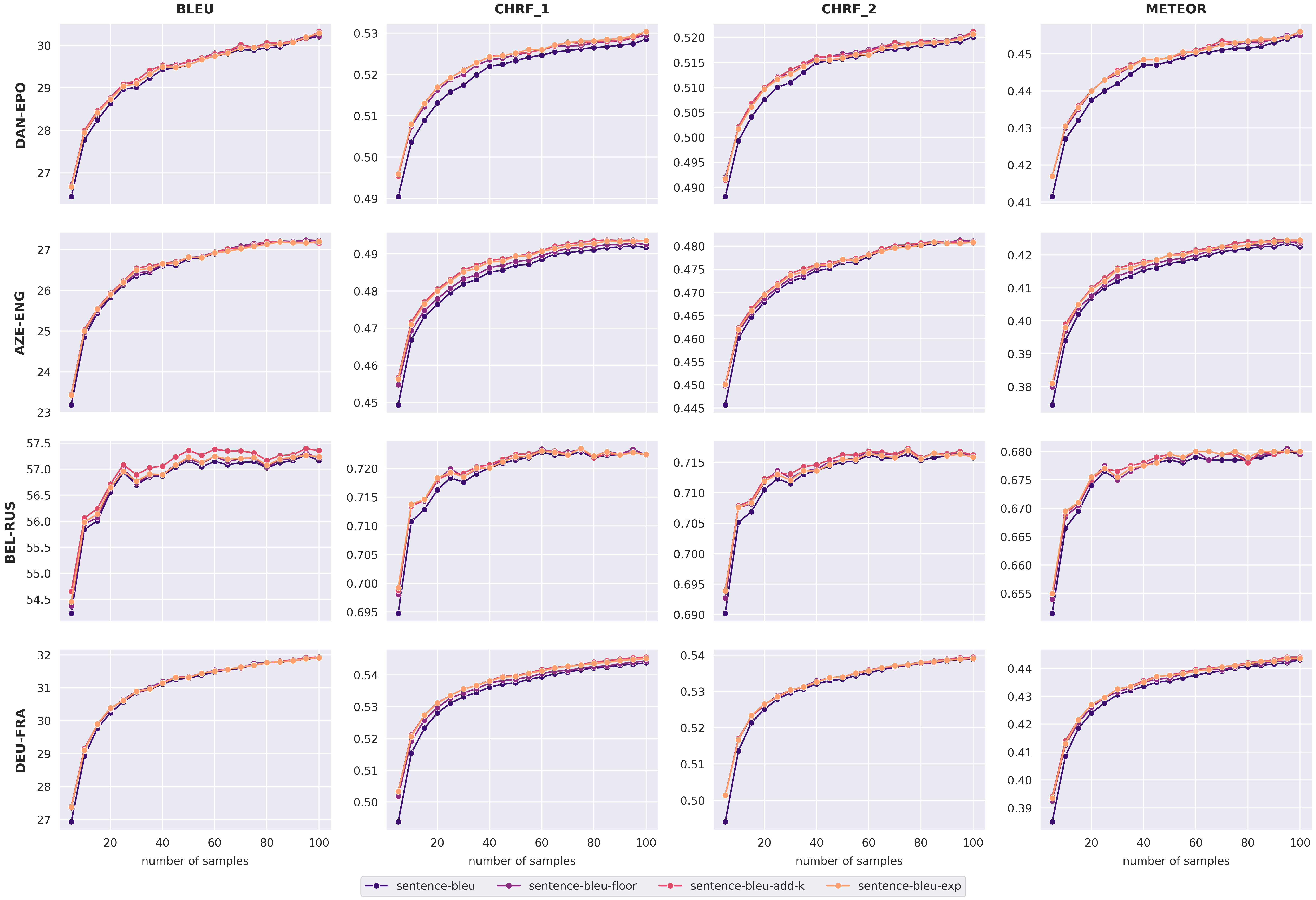}
    \caption{Comparison of utility functions that are variants of BLEU.}
    \label{fig:results_compare_risk_functions_bleu}
\end{figure*}

Figures \ref{fig:results_compare_risk_functions_chrf} and \ref{fig:results_compare_risk_functions_bleu} show additional results for MBR decoding with utility functions that are variants of CHRF and BLEU.


\section{Additional length tables}
\label{sec:additional-length-tables}

We provide additional length statistics for utility functions used with MBR in Table \ref{tab:additional-lengths}.


\begin{table*}
    \centering
    \begin{tabular}{lcccc}
      \toprule
       & \multicolumn{1}{c}{\textbf{DAN-EPO}} & \multicolumn{1}{c}{\textbf{AZE-ENG}} & \multicolumn{1}{c}{\textbf{BEL-RUS}} & \multicolumn{1}{c}{\textbf{DEU-FRA}} \\
      \midrule
      reference & 11.91 & 15.54 & 8.41 & 20.19 \\
      \midrule
      sample & 11.73 & 15.15 & 8.29 & 19.99 \\
      beam-normalized & 11.61 & 14.45 & 8.23 & 19.62 \\
      beam-unnormalized & 11.21 & 13.62 & 8.20 & 19.08 \\
      \midrule
      bleu & 11.54 & 14.45 & 8.17 & 19.59 \\
      bleu-floor & 11.51 & 14.41 & 8.18 & 19.55 \\
      bleu-add-k & 11.46 & 14.29 & 8.20 & 19.40 \\
      bleu-exp & 11.42 & 14.29 & 8.18 & 19.41 \\
      \midrule
      bleu-symmetric & 11.55 & 14.39 & 8.19 & 19.58 \\
      bleu-floor-symmetric & 11.51 & 14.34 & 8.19 & 19.53 \\
      bleu-add-k-symmetric & 11.39 & 14.14 & 8.19 & 19.25 \\
      bleu-exp-symmetric & 11.41 & 14.21 & 8.18 & 19.37 \\
      \midrule
      chrf-1 & 11.48 & 14.16 & 8.18 & 19.40 \\
      chrf-2 & 12.50 & 15.88 & 8.31 & 20.89 \\
      chrf-3 & 13.01 & 16.92 & 8.45 & 21.93 \\
      \midrule
      chrf-1-symmetric & 11.48 & 14.16 & 8.18 & 19.40 \\
      chrf-2-symmetric & 11.48 & 14.16 & 8.18 & 19.40 \\
      chrf-3-symmetric & 11.48 & 14.16 & 8.18 & 19.40 \\
      \midrule
      chrf-0.5 & 10.63 & 12.99 & 8.08 & 18.02 \\
      \midrule
      meteor & 12.23 & 15.29 & 8.26 & 20.38 \\
      meteor-symmetric & 11.47 & 14.12 & 8.20 & 19.40 \\
      \bottomrule
    \end{tabular}
    \caption{Lengths of hypotheses as mean number of tokens.}
    \label{tab:additional-lengths}
\end{table*}

\end{document}